\documentclass{article}

\usepackage{arxiv}
\usepackage{algorithm}
\usepackage{algpseudocode}
\usepackage{amsmath}

\usepackage{adjustbox}
\usepackage{array}
\usepackage{multirow}
\usepackage{caption}

\usepackage[utf8]{inputenc} 
\usepackage[T1]{fontenc}    
\usepackage{hyperref}       
\usepackage{url}            
\usepackage{booktabs}       
\usepackage{amsfonts}       
\usepackage{nicefrac}       
\usepackage{microtype}      
\usepackage{lipsum}
\usepackage{graphicx}
\graphicspath{ {./images/} }

\title{ANUBHUTI: A Comprehensive Corpus
 for Sentiment Analysis in Bangla
 Regional Languages}

\author{
Swastika Kundu\thanks{Corresponding author} \\
Department of Computer Science and Engineering\\
Ahsanullah University of Science and Technology\\
Dhaka, Bangladesh \\
\texttt{swastika.cse.20210104084@aust.edu}
\And
Autoshi Ibrahim \\
Department of Computer Science and Engineering\\
Ahsanullah University of Science and Technology\\
Dhaka, Bangladesh \\
\texttt{autoshi.cse.20210104010@aust.edu}
\And
Mithila Rahman \\
Department of Computer Science and Engineering\\
Ahsanullah University of Science and Technology\\
Dhaka, Bangladesh \\
\texttt{mithila.cse.20200204057@aust.edu}
\And
Tanvir Ahmed \\
Department of Computer Science and Engineering\\
Ahsanullah University of Science and Technology\\
Dhaka, Bangladesh \\
\texttt{tanvir.cse@aust.edu}
}

\def\bng{\bngx}

\font\bngx=bang10

\def\*#1*#2{o\null{#2}{#1}}

\def\sh#1{\setbox0=\hbox{#1}%
     \kern-.02em\copy0\kern-\wd0
     \kern.04em\copy0\kern-\wd0
     \kern-.02em\raise.0433em\box0 }
\begin{document}
\maketitle

\begin{abstract}

Sentiment analysis of Bangla's regional dialects is underexplored due to linguistic diversity and a scarcity of annotated data. This paper introduces ANUBHUTI, a comprehensive dataset consisting of 10,000 sentences manually translated from standard Bangla into four major regional dialects—Mymensingh, Noakhali, Sylhet, and Chittagong. The dataset predominantly features political and religious content—reflecting Bangladesh's contemporary socio-political landscape—alongside neutral texts for balance. Each sentence is annotated using a dual annotation scheme: (i) multiclass thematic labeling categorizes sentences as Political, Religious, or Neutral, and (ii) multilabel emotion annotation assigns one or more emotions from Anger, Contempt, Disgust, Enjoyment, Fear, Sadness, and Surprise. Expert native translators performed the translation and annotation, with quality assurance achieved via Cohen’s Kappa inter-annotator agreement, ensuring strong consistency across dialects. Systematic checks further refined the dataset for missing data, anomalies, and inconsistencies. ANUBHUTI fills a critical resource gap for sentiment analysis in low-resource Bangla dialects, enabling more accurate, context-aware natural language processing.

\end{abstract}

\keywords{Regional Languages \and Sentiment Analysis \and Corpus \and Dataset \and Natural Language Processing}



\section{Significance of ANUBHUTI}

\begin{itemize}
    \item ANUBHUTI is a pioneering corpus for sentiment analysis in Bangla's regional dialects, addressing a critical gap in low-resource NLP research for dialects like Sylhet, Chittagong, Noakhali, and Mymensingh.
    
    \item Expert linguists specializing in regional Bangla dialects thoroughly annotated the dataset, ensuring a high-quality resource for sentiment analysis.
    
    \item The corpus facilitates dialect-aware chatbots, mental health analysis, and social media monitoring, contributing to inclusive, regionally adaptive AI technologies.
    
    \item The dataset's structured CSV format ensures compatibility with various NLP pipelines, making it a valuable resource for academic and industry research in Bangla language processing.
\end{itemize}

\section{Related Works}
Sentiment and emotion analysis in morphologically rich, low-resource languages has grown considerably in recent years. However, progress has primarily focused on high-resource languages or standardized variants, leaving dialectal and regionally diverse linguistic forms underexplored. This literature review synthesizes recent studies across Bangla, Arabic, and Hindi, with a particular emphasis on dialect-aware sentiment classification, resource development, and methodological innovation.

\subsection{Bangla Sentiment and Emotion Analysis}
Bangla, as a low-resource yet widely spoken language, has increasingly become the focus of sentiment and emotion analysis efforts. Bhowmik et al. (2021)\cite{bhowmik2021bangla} proposed a supervised learning framework combining a categorical weighted lexicon (LDD) with a rule-based sentiment scoring method called Bangla Text Sentiment Score (BTSC). Their SVM model achieved 82.21\% classification accuracy on cricket and restaurant review datasets using BiGram features. However, the study underscored the shallow progress of Bangla sentiment analysis due to a lack of computational resources and annotated corpora. To address the need for large-scale data, Kabir et al. (2023)\cite{kabir2023banglabook} introduced BanglaBook, a dataset of over 158,000 Bangla book reviews categorized into positive, neutral, and negative sentiments. The authors evaluated multiple traditional and deep learning models, including fine-tuned Bangla-BERT, which achieved a weighted F1-score of 0.933. The work emphasized the shortage of domain-specific annotated datasets and advanced pre-trained models tailored for Bangla.

In a machine-learning-focused study, Paul et al. (2025)\cite{paul2025analyzing} analyzed Bangla social media comments using ML classifiers, BiLSTM, and LIME for explainability. Leveraging the EmoNoBa dataset with six emotion classes, their Decision Tree model boosted with AdaBoost achieved an F1-score of 0.786. The study addressed concerns that social media datasets may not fully reflect Bangla linguistic usage. In another recent study, Mahmud et al.(2024)\cite{mahmud2024benchmark} presented a benchmark dataset for sentiment analysis of Bangla cricket-related social media content, focusing on low-resource and noisy textual environments. With a Cohen’s Kappa score of 0.971 indicating strong annotation reliability, the dataset supported evaluations using deep learning models. The BiLSTM model achieved 95.2\% accuracy, outperforming both LSTM and traditional RNN models. This work addressed the scarcity of annotated Bangla datasets in the sports domain. Islam et al. (2023)\cite{islam2023sentigold} further advanced resource development by releasing SentiGOLD, a 70,000-sample Bangla dataset spanning over 30 domains such as news, blogs, and social media. Featuring a nuanced five-class sentiment scale, SentiGOLD achieved an Inter-Annotator Agreement of 0.88. Experiments showed that BanglaBERT performed competitively both in in-domain classification and zero-shot cross-dataset evaluations, outperforming prior datasets like SentNoB. The authors addressed issues of scale, bias, and annotation quality prevalent in earlier Bangla resources.



\subsection{Sentiment Analysis in Arabic Dialects}
A foundational contribution in dialectal Arabic sentiment analysis is the work by Baly et al. (2017)\cite{baly2017comparative}, who introduced the Multi-Dialect Arabic Sentiment Twitter Dataset (MD-ArSenTD), containing tweets from 12 Arab countries with a focus on Egyptian and UAE dialects. The study compared Support Vector Machines (SVM) with engineered features against deep learning models like Long Short-Term Memory (LSTM). Results showed that LSTM models, when supplemented with dialect-specific word embeddings, outperformed traditional methods. Egyptian tweets, being highly dialectal, benefited more from morphological and localized lexical features, while UAE tweets were more aligned with Modern Standard Arabic (MSA). The work was the first to conduct sentiment analysis on UAE tweets and applied a novel five-point sentiment scale for Egyptian tweets.

Similarly, Mdhaffar et al. (2017)\cite{mdhaffar2017sentiment} tackled sentiment analysis for the Tunisian dialect, creating the first publicly available annotated sentiment corpus for Tunisian Arabic (TSAC), consisting of 17,000 Facebook comments. They experimented with classifiers including MLP, Naive Bayes, and SVM, and demonstrated that models trained on dialect-specific data yielded significantly better performance than those trained on other dialects or MSA. The study emphasized the need for tailored linguistic resources and highlighted the scarcity of NLP tools for North African Arabic dialects.

\subsection{Sentiment Analysis in Hindi}
In the context of Hindi, Pandey and Govilkar (2015)\cite{pandey2015framework} proposed an unsupervised sentiment analysis framework leveraging HindiSentiWordNet (HSWN). While effective for basic polarity detection, the HSWN lexicon was limited in coverage—particularly in handling adjectives and adverbs—and struggled with context-dependent word interpretation. The paper noted that without sufficient lexical expansion and automated context handling, accurate sentiment extraction in Hindi remains challenging.

\subsection{Dialect-Based Sentiment Analysis}
A major enabling resource for dialectal Bangla NLP is Vashantor, proposed by Faria et al. (2023)\cite{faria2023vashantor}. This benchmark dataset facilitates the automated translation of regional Bangla dialects into standard Bangla. Covering a wide geographic and linguistic span, including Sylhet, Noakhali, Barishal, Chittagong, and Mymensingh dialects, Vashantor is a pivotal contribution to dialect normalization and low-resource translation. Although not focused directly on sentiment analysis, the dataset provides foundational infrastructure to build dialect-aware NLP models and enhance downstream tasks such as emotion or sentiment classification in regional varieties of Bangla. Similar to Vashantor dataset, Sultana et al. (2025)\cite{sultana2025onubad} and Paul et al. (2025)\cite{paul2025ancholik} proposed their own version of Dialect Translation dataset. In a recent study of Regional Dialect, Paul et al. (2024)\cite{paul2024improving} explored regional dialect classification using data from the VASANTOR dataset, covering Bangla dialects like Mymensingh, Chittagong, Barishal, Noakhali, and Sylhet. The study compared BanglaBERT, GPT-3.5 Turbo, and Gemini 1.5 Pro, finding that BanglaBERT achieved the highest classification accuracy at 88.74\%. This work underlines the complexity of intra-language variation and the need for dialect-sensitive models.

Despite significant advancements in sentiment and emotion analysis for standard Bangla, Hindi, and Arabic dialects, a notable research gap persists in the area of regional dialect sentiment analysis, particularly for low-resource languages like Bangla. Existing studies either rely heavily on standard language forms or lack the granularity to capture regional linguistic nuances. While efforts like Vashantor and regional classification models have laid the groundwork, sentiment-specific datasets and models for Bangla dialects remain virtually nonexistent. This underscores the urgent need for sentiment analysis systems that understand and process non-standard dialects, prevalent in real-world communication but underrepresented in current NLP infrastructure. In response to this gap, our research aims to advance dialect-aware sentiment analysis by leveraging modern transformer architectures, domain adaptation, and interpretability techniques—thus contributing to a more inclusive and linguistically representative natural language processing landscape.

\section{Data Description}
Sentiment analysis, also known as opinion mining, is a fundamental task in Natural Language Processing (NLP) that involves identifying and categorizing emotions or opinions expressed in text. It is vital in applications such as social media analysis, political forecasting, and consumer behavior research, by assessing the polarity of textual input.

While sentiment analysis has progressed substantially in high-resource languages like English, its development in low-resource languages such as Bangla remains nascent. Most Bangla sentiment analysis focuses on formal or standard language, overlooking the rich diversity of dialectal variations across Bangladesh. However, dialectal expressions are often more emotionally charged, contextually grounded, and reflective of grassroots-level sentiment. Addressing this gap, our work focuses on sentiment analysis in Bangla's regional dialects. This approach enables a more inclusive, granular understanding of public opinion, especially in Bangladesh's socio-political and religious online discourse.

\begin{figure}[h]
    \centering
    \includegraphics[width=0.8\textwidth]{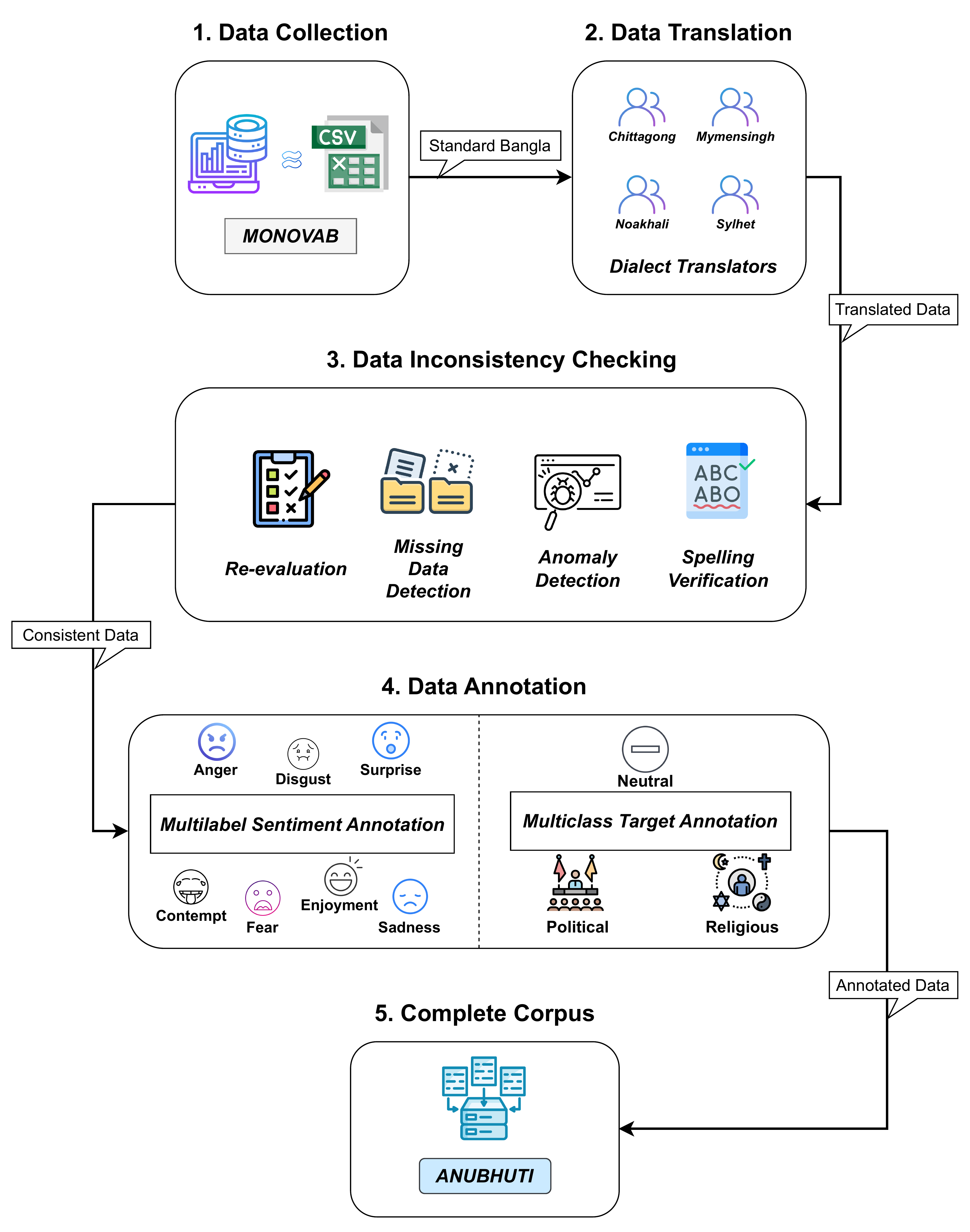} 
    \caption{Systematic pipeline for the development of the ANUBHUTI dataset, illustrating data collection, preprocessing, and annotation stages.}
    \label{fig:Data_Pipeline}
\end{figure}

To support this goal, we developed a dialect-sensitive sentiment dataset named \textbf{ANUBHUTI}, which is tailored to capture the linguistic and emotional nuances present in regional Bangla. Figure~\ref{fig:Data_Pipeline} illustrates the development pipeline of the ANUBHUTI dataset, highlighting the key stages of data sourcing, filtering, preprocessing, and annotation.

\subsection{Data Collection}
The dataset used in this study, titled ANUBHUTI, was developed using raw data sourced from the publicly available \href{https://figshare.com/articles/dataset/_MONOVAB_An_Annotated_Corpus_for_Bangla_Multi-label_Emotion_Detection/24199260?file=42467856}{MONOVAB} dataset. MONOVAB comprises a diverse range of user-generated Bangla content, making it a valuable resource for mining sentiment-rich expressions in regional dialects.

Initially, we collected 2,500 raw data samples for preliminary sentiment analysis. The primary focus was on political and religious discourse, as these domains tend to evoke strong emotional reactions and diverse public opinions. A subset of neutral content was also included to provide a baseline for comparative sentiment evaluation. The choice of political and religious content reflects Bangladesh's current socio-political climate, where heightened public discourse around governance, ideological tensions, and religious sentiments is prominent on digital platforms. These topics often reflect deeper emotional undercurrents and are highly relevant for training sentiment classification models that aim to capture context-sensitive regional expressions.

\subsection{Data Translation}
To construct a regionally diverse sentiment dataset, we selected four prominent dialect regions of Bangladesh: Mymensingh, Noakhali, Sylhet, and Chittagong. These regions were chosen based on their distinct linguistic features and strong presence in local digital discourse. To ensure balanced representation, we translated 2,500 sentences into the corresponding dialect for each region, resulting in a total of 10,000 dialectal samples.

Due to the complexity and variability of regional expressions, we hired two native dialect experts for each region to perform the translation. These experts' linguistic familiarity and contextual understanding of local expressions allowed them to preserve each sentence's semantic integrity and sentiment polarity during translation. The paired experts' collaboration ensured consistency, resolved ambiguities, and maintained dialectal authenticity. This translation was crucial for adapting sentiment-labeled content into natural, expressive dialectal forms reflecting real-world language use. The translated data serves as a foundation for training sentiment analysis models that can operate effectively on informal and region-specific Bangla text.

\subsection{Data Translator Profiles}
To ensure linguistic and contextual accuracy, we recruited eight native speakers with proven expertise in their regional dialects—two for each region: Mymensingh, Noakhali, Sylhet, and Chittagong. All translators were either language professionals, postgraduate students in linguistics, or individuals with prior experience in language documentation or translation work. Their familiarity with regional speech patterns, idiomatic expressions, and sentiment articulation made them suitable for the task. A thorough screening process was conducted to assess their dialectal proficiency and ensure consistency in translation quality. Detailed information about the translators is presented in Table~\ref{tab:translator_info}.  

\begin{table}[h]
\centering
\caption{Background and expertise of dialect translators, including their native languages, years of experience, and familiarity with specific dialects}
\label{tab:translator_info}
\begin{tabular}{|c|c|c|c|}
\hline
\textbf{Region} & \textbf{ID} & \textbf{Background} & \textbf{Expertise} \\
\hline
Mymensingh & T1, T2 & Under GraduateStudent & Native Speaker, Translation Experience \\
Noakhali   & T3, T4 & GraduateStudent, Journalist & Native Speaker \\
Sylhet     & T5, T6 & Researcher & Native Speaker \\
Chittagong & T7, T8 &  UnderGraduate & Native Speaker \\
\hline
\end{tabular}
\end{table}

\subsection{Cohen’s Kappa for Translation Agreement}

To ensure the quality and reliability of ANUBHUTI's dialectal translations, we evaluated inter-annotator agreement between the two translators assigned to each region. Given the subjective nature of language translation, especially in dialects without a standardized written form, a robust agreement metric was essential. 

We used Cohen’s Kappa ($\kappa$) to quantify the level of agreement between the two translators, beyond chance. Unlike simple percentage agreement, Cohen's Kappa accounts for agreement occurring by chance, making it a more reliable metric for subjective tasks like translation validation. A high $\kappa$ score indicates strong consistency and shared understanding between the translators, which is critical for maintaining sentiment and semantic fidelity in the dataset. As shown in Table~\ref{tab:kappa_scores}, the scores for all four regions—Mymensingh, Noakhali, Sylhet, and Chittagong—fall within the range of 0.76 to 0.84, indicating \textit{substantial to almost perfect agreement}.

The computation was based on a comparison between the annotations provided by the two translators per region, using the standard Cohen’s Kappa formula. Agreement was recorded for each sentence if both translators provided semantically and sentimentally aligned translations. The observed agreement ($p_o$) and the expected agreement by chance ($p_e$) were computed from the annotation distribution, and Cohen's Kappa ($\kappa$) was derived using:

\[
\kappa = \frac{p_o - p_e}{1 - p_e}
\]

Where:
\begin{itemize}
    \item $p_o$ is the observed agreement between annotators,
    \item $p_e$ is the expected agreement by chance.
\end{itemize}

\begin{table}[h]
\centering
\caption{Cohen's Kappa scores quantifying inter-translator agreement for dialect translation, grouped by geographic region.}
\label{tab:kappa_scores}
\begin{tabular}{|c|c|}
\hline
\textbf{Region} & \textbf{Kappa Score ($\kappa$)} \\
\hline
Mymensingh & 0.84 \\
Noakhali   & 0.78 \\
Sylhet     & 0.81 \\
Chittagong & 0.76 \\
\hline
\end{tabular}
\end{table}

These region-specific scores confirm a high degree of consistency and mutual understanding between translators, reinforcing the quality and reliability of the ANUBHUTI dataset’s dialectal representations.

\subsection{Data Quality Assurance}

Ensuring consistency and quality across all translated data is crucial for this sentiment analysis research on Bangla's regional dialects.Because the original dataset in standard Bangla was manually translated into four regional dialects—Mymensingh, Noakhali, Sylhet, and Chittagong—the risk of inconsistency was high due to linguistic variability, subjective translation choices, and human error. To address these challenges, a systematic and multi-stage \textit{data inconsistency checking framework} was developed. This framework consists of four primary stages:

\subsubsection{Semantic and Sentiment Verification}
In this phase, all translated sentences were manually reviewed and cross-checked against the original standard Bangla texts. The goal was to ensure semantic alignment and sentiment preservation. Reviewers focused on idiomatic correctness, dialect-specific vocabulary, and the emotional tone of each sentence. This process helped guarantee that the translated dialectal texts accurately reflected the original content's intent and polarity.

\subsubsection{Missing Data Detection}

To ensure completeness and reliability, we used a dual-phase approach to identify missing or incomplete translations. This involved both manual inspection and automated detection using Python scripting.

The manual review process was conducted to check for:
\begin{itemize}
    \item Entirely absent translations,
    \item Partially completed sentences,
    \item Omission of key sentiment-bearing words or phrases.
\end{itemize}

Each translated entry was validated against its original Bangla source to ensure semantic completeness and fidelity of sentiment expression, reducing the risk of information loss during translation.

To complement this, a Python-based script was developed to automatically identify missing values within the dataset. The script reads the dataset using the \texttt{pandas} library, checks each column for \texttt{NaN} or null values, and summarizes the missing data count. If any missing entries are detected, the script provides a detailed report for further review. The core logic of this automated detection system is outlined in \textbf{Algorithm~\ref{alg:missing_data}}.

\begin{algorithm}
\caption{Missing data check results in the ANUBHUTI dataset CSV file, displaying the count and percentage of missing values for each column}
\label{alg:missing_data}
\begin{algorithmic}[1]
\Procedure{CheckMissingData}{filepath}
    \State Import the \texttt{pandas} library
    \State \texttt{df} $\gets$ \texttt{pandas.read\_csv(filepath)}
    \State \texttt{missing\_values} $\gets$ \texttt{df.isnull().sum()}
    \State \texttt{total\_missing} $\gets$ \texttt{df.isnull().sum().sum()}
    \If{\texttt{total\_missing > 0}}
        \State Print ``Missing values found:''
        \State Print \texttt{missing\_values}
    \Else
        \State Print ``No missing values found.''
    \EndIf
\EndProcedure
\end{algorithmic}
\end{algorithm}

\subsubsection{Anomaly Detection}

To ensure the quality and dialectal integrity of the translated dataset, an anomaly detection phase was implemented. This phase aimed to identify irregularities that deviated from the expected linguistic and contextual norms of each target dialect. Anomalies were broadly categorized into two types: \textit{labeling inconsistencies} and \textit{textual anomalies}.

Manual review was used to detect issues such as:
\begin{itemize}
    \item Sentences that mixed multiple dialects or included standard Bangla unintentionally,
    \item Translations that were contextually inappropriate or semantically misaligned with the original sentence,
    \item Use of grammatical constructions that did not conform to the syntax of the target dialect.
\end{itemize}

In addition to manual checking, we developed a semi-automated Python script to systematically flag anomalous entries. The script evaluated sentence formatting to identify unusual patterns such as excessive punctuation (e.g., ``!!!!'', ``....''), non-Bangla characters or scripts, and other spam-like structures. Such anomalies were considered indicators of noise or stylistic inconsistency that could impact downstream analysis.

All automatically flagged entries were reviewed manually to confirm their anomalous status, and necessary corrections were applied to preserve semantic, emotional, and dialectal integrity. The structured logic of this anomaly detection procedure is presented in \textbf{Algorithm~\ref{alg:anomaly_detection}}.

\begin{algorithm}
\caption{Anomaly detection results in the Bangla sentiment dataset, highlighting data points identified as outliers}
\label{alg:anomaly_detection}
\begin{algorithmic}[1]
\Procedure{DetectAnomalies}{Dataset}
    \State Import \texttt{pandas, re, sentence-transformers, sklearn, fuzzywuzzy}
    \State \texttt{df} $\gets$ Load the CSV file
    
    \Statex
    \For{each text in \texttt{df[`text']}}
        \If{Contains repeated chars (e.g., ``!!!!'' or ``....'') or non-Bangla scripts}
            \State Flag as \texttt{Content Anomaly: Suspect formatting or language}
        \EndIf
    \EndFor
\EndProcedure
\end{algorithmic}
\end{algorithm}

\subsubsection{Dialectal Spelling Standardization}
Due to the lack of standardized orthography for many regional Bangla dialects, spelling verification was performed manually with the aid of native speakers. The goal was to ensure:
\begin{itemize}
    \item Consistency in commonly accepted dialectal spellings,
    \item Phonetic alignment with regional pronunciation,
    \item Elimination of ambiguities caused by irregular spelling variants.
\end{itemize}
Given the informal nature of written regional dialects, this step was essential for preserving clarity, authenticity, and interpretability in the translated texts.

\vspace{0.5em}
\noindent The manual inconsistency checking process played a foundational role in enhancing ANUBHUTI's reliability and validity. By ensuring translation accuracy, completeness, contextual consistency, and dialect-specific spelling correctness, the dataset was rendered suitable for training robust and dialect-aware sentiment analysis models. This quality control phase ensures that models can effectively interpret and classify emotional content in diverse forms of Bangla, particularly from underrepresented regional communities.

\subsection{Data Annotation}

To prepare the ANUBHUTI dataset for supervised sentiment analysis, a two-level annotation process was carried out. The annotation involved manually labeling each sentence for its thematic class (multiclass) and emotional sentiment (multilabel). These annotations enable developing and evaluating models capable of understanding complex emotional expression and contextual bias in Bangla's regional dialects. The decision to implement both multiclass and multilabel annotation schemes was motivated by the need to capture the full complexity of sentiment expression in regional Bangla dialects. The multiclass annotation provides a high-level thematic categorization—distinguishing whether a sentence is political, religious, or neutral—allowing for context-aware analysis of sentiment trends. Meanwhile, the multilabel emotion tagging captures the diverse and overlapping emotional states conveyed in informal dialectal speech.

\subsubsection{Thematic Label Assignment}

Each sentence in the dataset was annotated with one of three mutually exclusive thematic labels: \texttt{Political}, \texttt{Religious}, or \texttt{Neutral}. This multiclass classification captures the overarching context or topic of the sentence, and is crucial for understanding domain-specific sentiment patterns.

\begin{itemize}
    \item \texttt{Political:} Includes references to governance, elections, political figures, movements, or actions.
    \item \texttt{Religious:} Encompasses content related to faith, belief systems, religious practices, and figures.
    \item \texttt{Neutral:} Covers general content without explicit political or religious context.
\end{itemize}

Each sentence was assigned exactly one of these classes, ensuring that thematic context is represented clearly and non-overlappingly. This classification supports topic-aware sentiment analysis and can be used to analyze how sentiment varies across different socio-cultural domains.

\subsubsection{Emotional Label Assignment}

In addition to contextual classification, each sentence was manually annotated for emotional sentiment using a multilabel scheme. The following seven core emotions were used as annotation labels:

\begin{itemize}
    \item \texttt{Anger:} Assigned to sentences that express frustration, outrage, or hostility, often directed at political figures, events, or institutions. For example, accusatory or confrontational statements about government actions or religious intolerance.
    \item \texttt{Contempt:} Applied to sarcastic, belittling, or demeaning expressions — particularly those mocking political leaders, religious opponents, or ideologies. Contemptuous tone is often implicit and requires contextual understanding.
    \item \texttt{Disgust:} Used for sentences reflecting moral revulsion, societal disappointment, or emotional rejection. This often includes strong criticisms of corruption, injustice, or offensive actions within political or religious contexts.
    \item \texttt{Enjoyment:} Assigned to sentences expressing positive emotion, such as approval, support, or satisfaction. In political or religious contexts, this might include praise for actions, victories, or celebrations of community events.
    \item \texttt{Fear:} Used for sentences that indicate anxiety, insecurity, or anticipation of negative consequences. This often includes references to political instability, religious violence, or threats to personal or communal safety.
    \item \texttt{Sadness:} Assigned when the sentence reflects grief, loss, or sorrow — such as mentions of deaths, tragedies, or despair related to social or political events.
    \item \texttt{Surprise:} Applied to sentences indicating unexpected outcomes or shock — whether positive or negative. This may involve disbelief at political scandals, sudden policy changes, or religious controversies.
\end{itemize}

A sentence could exhibit one, multiple, or even no emotion labels depending on its emotional tone. This multilabel approach captures the nuance and complexity of real-life emotional expression, especially within dialectal text where mixed sentiment is common. For example, a politically charged sentence might simultaneously express both \texttt{Anger} and \texttt{Contempt}, while another might express only \texttt{Sadness}. Annotators were instructed to label all applicable emotions to ensure comprehensive emotional coverage.

\subsubsection{Annotation Process}

The annotation of the ANUBHUTI dataset was carried out manually by the same regional dialect experts who were previously involved in the translation of the dataset. Their linguistic familiarity and contextual awareness made these translators well-positioned to interpret the texts' literal meaning and emotional undertone within their dialectal context.

Each translator team was responsible for annotating the 500 dialectal sentences corresponding to their assigned region: Mymensingh, Noakhali, Sylhet, and Chittagong. They performed both levels of annotation:

\begin{itemize}
    \item \textbf{Multiclass Annotation:} Assigning one thematic category (Political, Religious, or Neutral) per sentence based on its subject matter.
    \item \textbf{Multilabel Annotation:} Selecting one or more emotion labels from a predefined set of seven (Anger, Contempt, Disgust, Enjoyment, Fear, Sadness, Surprise), according to the emotional tone conveyed in each sentence.
\end{itemize}

\begin{table}[h!]
\centering
\caption{Overview of dataset annotation process for the Chittagong region, detailing the annotation guidelines, tools used, and quality control measures}
\label{tab:emotion_target}
\begin{tabular}{|p{2.5cm}|c|c|c|c|c|c|c|c|}
\hline
\textbf{Chittagong} & \textbf{Target} & \textbf{anger} & \textbf{contempt} & \textbf{disgust} & \textbf{enjoyment} & \textbf{fear} & \textbf{sadness} & \textbf{surprise} \\
\hline
{\bng inU maer/kT Ar sNNGghr/Shr ghTna {I}yaner{O} raj{oi}nitk ruup ed{O}yn Ar la{I} srkarer Aibhnn/dn.ibim/pr entaHr/mii ebyag/Yun Aasaim.} & Political & 1 & 1 & 0 & 0 & 0 & 0 & 0 \\ \hline
{\bng Engir Aa{I}nr shaShn pRitSh/Tha grn ed?} &Political & 0 & 1 & 1 & 0 & 0 & 0 & 0 \\ \hline
{\bng Aar HNNidn naTk etara girib} &Neutral   & 1 & 0 & 1 & 0 & 0 & 0 & 0 \\ \hline

\hline
\end{tabular}
\end{table}
Table~\ref{tab:emotion_target} presents a sample of the final annotation format used in the ANUBHUTI dataset. Each row corresponds to a single sentence translated into the Chittagong dialect and annotated with both a thematic target class (Political, Religious, or Neutral) and multilabel emotional categories. The emotion labels—Anger, Contempt, Disgust, Enjoyment, Fear, Sadness, and Surprise—are represented using binary indicators (1 for presence, 0 for absence), allowing for the expression of multiple emotions per sentence. While the table displays data from only the Chittagong region for illustration, the same annotation schema and format were applied consistently across all four regions: Mymensingh, Noakhali, Sylhet, and Chittagong. This uniform structure ensures compatibility for cross-regional sentiment analysis and enables meaningful comparative modeling.

\begin{figure}[h]
    \centering
    \includegraphics[width=0.6\textwidth]{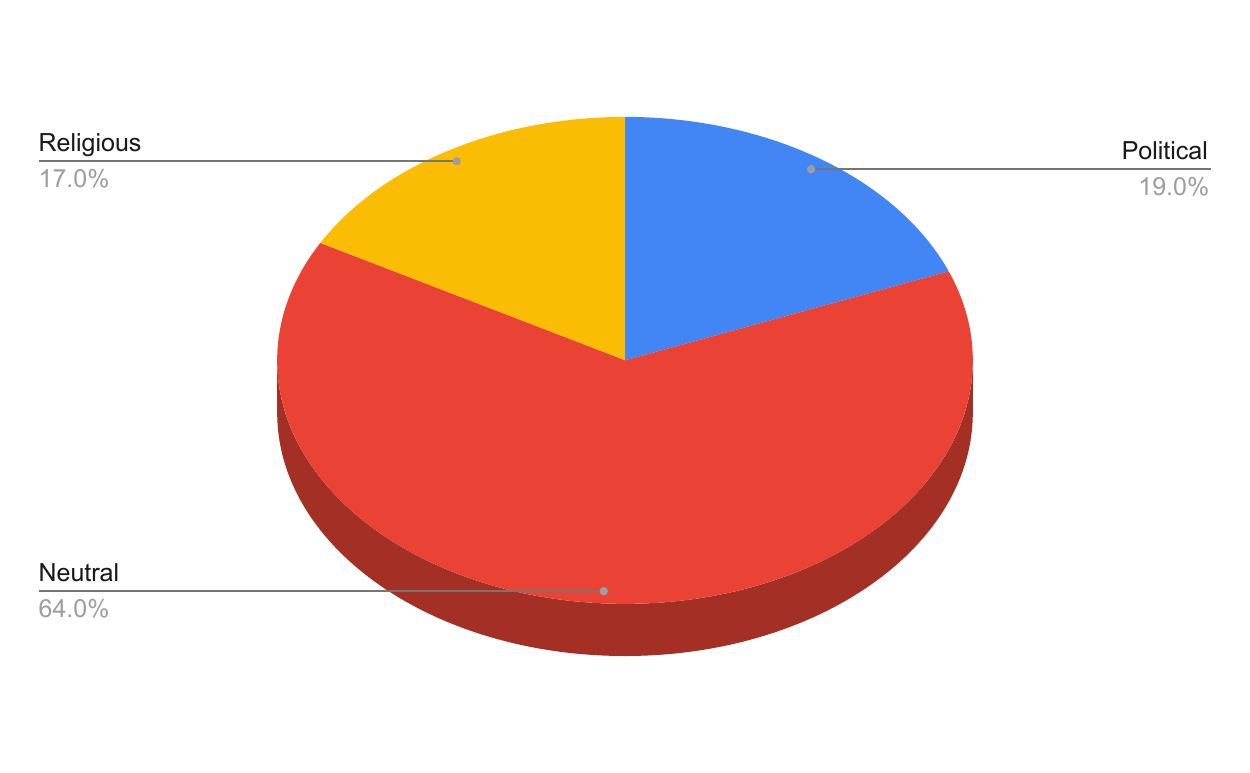} 
    \caption{Multiclass distribution of the ANUBHUTI dataset, showing the number of samples for each class label}
    \label{fig:Multiclass}
\end{figure}

\subsection{Dataset Statistics}

\begin{table}[]
\centering
\caption{Number of Sentences per Region in the ANUBHUTI Dataset}
\label{tab:sentence_distribution}
\begin{tabular}{|l|c|}
\hline
\textbf{Region} & \textbf{Number of Sentences} \\
\hline
Mymensingh & 2500 \\
Noakhali   & 2500 \\
Sylhet     & 2500 \\
Chittagong & 2500 \\
\hline
\textbf{Total} & \textbf{10,000} \\
\hline
\end{tabular}
\end{table}

The ANUBHUTI dataset comprises a total of 10,000 sentences distributed equally across four major regional dialects of Bangla. As shown in Table~\ref{tab:sentence_distribution}, each region—Mymensingh, Noakhali, Sylhet, and Chittagong—contributes 2500 sentences to ensure a balanced and comparable representation of dialectal variations. This equal allocation supports robust dialect-specific analysis and allows for fair model training and evaluation across regions. The dataset’s regional symmetry was a key design choice to avoid dialectal bias and to ensure methodological consistency in downstream sentiment analysis tasks.

\begin{figure}[h]
    \centering
    \includegraphics[width=0.8\textwidth]{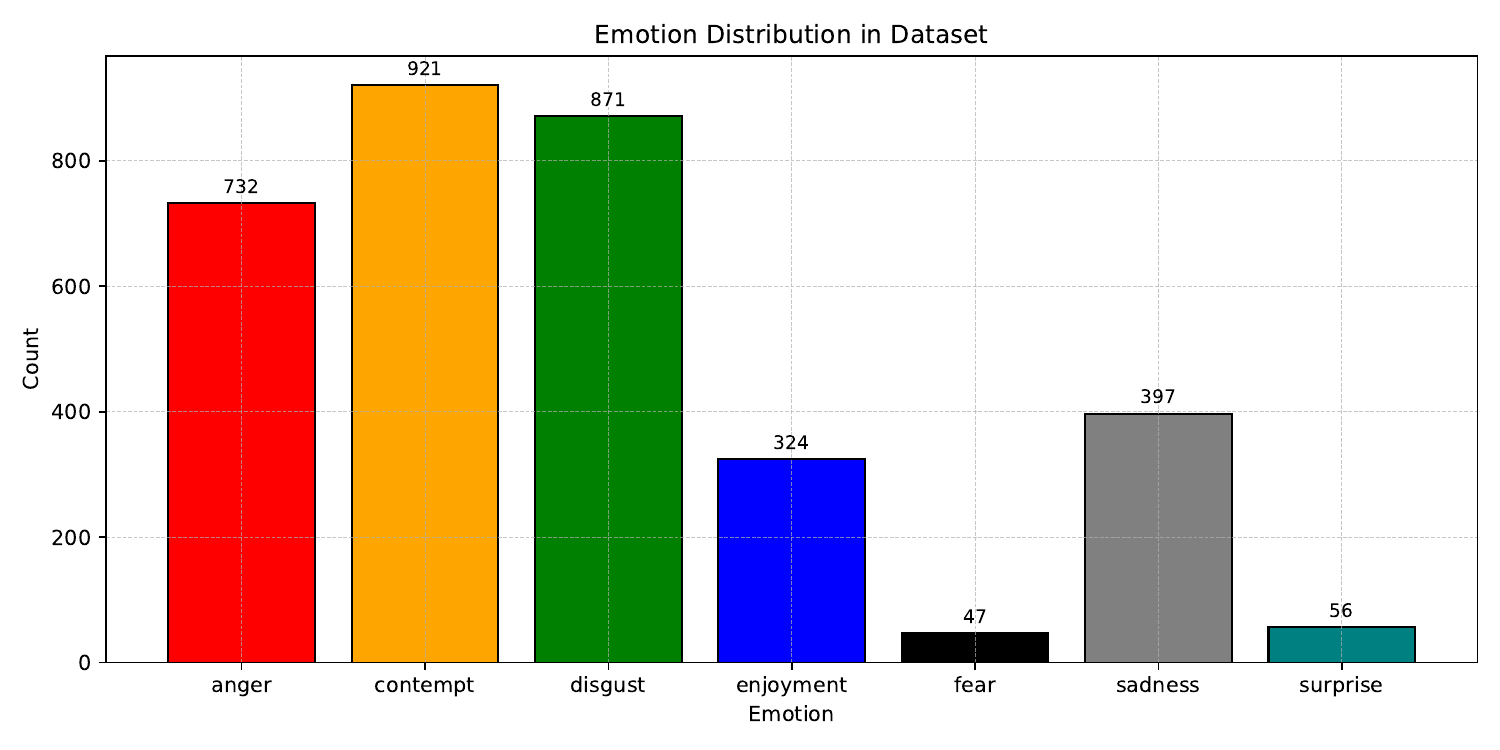} 
    \caption{Multilabel distribution of the ANUBHUTI dataset, displaying the frequency of each label combination within the dataset}
    \label{fig:Multilabel}
\end{figure}

Figure~\ref{fig:Multiclass} presents the thematic classification of sentences into three mutually exclusive categories: Political, Religious, and Neutral. Figure~\ref{fig:Multilabel} depicts the frequency of emotion labels assigned in the multilabel annotation scheme. The chart reveals that emotions such as Anger, Contempt, and Disgust occur more frequently, indicating the emotionally charged nature of the regional dialect texts, especially in political and religious discussions. Conversely, less frequent labels like Fear and Surprise reflect more nuanced or less commonly expressed sentiments.

Together, these figures provide a comprehensive overview of the dataset’s thematic and emotional composition, which is critical for designing and evaluating sentiment analysis models tailored to regional Bangla dialects.

\section{Limitations and Future Works}

Despite its contributions, the ANUBHUTI dataset has several limitations that should be acknowledged:

\begin{itemize}
    \item Limited Domain Coverage: The dataset primarily includes religious and political texts, which may not generalize well to other domains such as sports, entertainment, or daily conversation.

    \item Sentiment Class Imbalance: While efforts were made to balance sentiment classes, certain dialects may still exhibit skewed sentiment distributions, which can affect model fairness and generalization.
\end{itemize}

These limitations open directions for future improvements, such as collecting more naturally occurring dialectal text, expanding the dataset to include diverse domains, and increasing the number of samples per region.

\section{Conclusion}
In this study, we introduced ANUBHUTI, the first comprehensive sentiment analysis corpus tailored to Bangla regional dialects—namely Mymensingh, Noakhali, Sylhet, and Chittagong. By manually translating and annotating 2,500 dialect-specific sentences with both thematic and emotional labels, we addressed a critical gap in low-resource NLP for Bangla. The dual-layer annotation schema—combining multiclass contextual categorization and multilabel emotional sentiment—enables a richer, more nuanced understanding of public discourse, especially in politically and religiously sensitive domains. Our methodical approach to data quality, including native translator involvement, Cohen’s Kappa-based agreement validation, and multi-stage inconsistency checks, ensures that ANUBHUTI stands as a reliable resource for dialect-aware sentiment modeling. The balanced representation across dialects and the inclusion of diverse emotions empower researchers to develop robust, culturally grounded NLP systems capable of interpreting underrepresented linguistic communities.

Looking ahead, ANUBHUTI lays the foundation for building sentiment-aware chatbots, misinformation detectors, and mental health assessment tools in regional Bangla. We envision future expansions that include more dialects, broader topical coverage, and real-world applications. By releasing this dataset publicly, we aim to catalyze inclusive, dialect-sensitive NLP research and promote linguistic equity within Bangla language technology development.

\section{Ethics Statements}
The data used to construct ANUBHUTI does not raise ethical concerns, as it was collected from a publicly available dataset and through manual translation. The dataset contains no sensitive or private information, ensuring compliance with ethical standards. Additionally, no data was sourced from personal communications or restricted sources. This dataset is intended solely for NLP research and development; no human or animal subjects were involved in its creation.

\section{Availability of ANUBHUTI}
The ANUBHUTI dataset, containing annotated sentiment analysis data in Bangla regional dialects, is publicly available for research and academic purposes. Researchers interested in utilizing the dataset can access it through the following link:\href{https://data.mendeley.com/datasets/mjxwby94yw/3}{Dataset of Sentiment Analysis for Regional Bangla Language (Original data)} (Mendeley Data)

\section{CRediT Author Statement}

 \textbf{Swastika Kundu:} Data curation, Writing– original draft, Visualization, Validation \\ 
 \textbf{Autoshi Ibrahim:} Data curation, Investigation, Writing– original draft \\
 \textbf{Mithila Rahman:} Validation, Software, Conceptualization \\ 
 \textbf{Tanvir Ahmed:} Supervision, Writing– review \& editing

\section{Competing Interests Statement}

The authors declare that they have no known competing financial interests or personal relationships that could have appeared to influence the work reported in this paper.

\bibliographystyle{unsrt}  
\bibliography{references}  

@article{baly2017comparative,
  title={Comparative evaluation of sentiment analysis methods across Arabic dialects},
  author={Baly, Ramy and El-Khoury, Georges and Moukalled, Rawan and Aoun, Rita and Hajj, Hazem and Shaban, Khaled Bashir and El-Hajj, Wassim},
  journal={Procedia Computer Science},
  volume={117},
  pages={266--273},
  year={2017},
  publisher={Elsevier}
}

@inproceedings{mdhaffar2017sentiment,
  title={Sentiment analysis of tunisian dialects: Linguistic ressources and experiments},
  author={Mdhaffar, Salima and Bougares, Fethi and Esteve, Yannick and Hadrich-Belguith, Lamia},
  booktitle={Third Arabic natural language processing workshop (WANLP)},
  pages={55--61},
  year={2017}
}

@article{faria2023vashantor,
  title={Vashantor: a large-scale multilingual benchmark dataset for automated translation of bangla regional dialects to bangla language},
  author={Faria, Fatema Tuj Johora and Moin, Mukaffi Bin and Wase, Ahmed Al and Ahmmed, Mehidi and Sani, Md Rabius and Muhammad, Tashreef},
  journal={arXiv preprint arXiv:2311.11142},
  year={2023}
}

@article{pandey2015framework,
  title={A framework for sentiment analysis in Hindi using HSWN},
  author={Pandey, Pooja and Govilkar, Sharvari},
  journal={International Journal of Computer Applications},
  volume={119},
  number={19},
  year={2015},
  publisher={Citeseer}
}

@article{bhowmik2021bangla,
  title={Bangla text sentiment analysis using supervised machine learning with extended lexicon dictionary},
  author={Bhowmik, Nitish Ranjan and Arifuzzaman, Mohammad and Mondal, M Rubaiyat Hossain and Islam, Md S},
  journal={Natural Language Processing Research},
  volume={1},
  number={3},
  pages={34--45},
  year={2021},
  publisher={Springer}
}

@article{kabir2023banglabook,
  title={BanglaBook: A large-scale Bangla dataset for sentiment analysis from book reviews},
  author={Kabir, Mohsinul and Mahfuz, Obayed Bin and Raiyan, Syed Rifat and Mahmud, Hasan and Hasan, Md Kamrul},
  journal={arXiv preprint arXiv:2305.06595},
  year={2023}
}

@article{mahmud2024benchmark,
  title={A benchmark dataset for cricket sentiment analysis in bangla social media text},
  author={Mahmud, Tanjim and Karim, Rezaul and Chakma, Rishita and Chowdhury, Tanjia and Hossain, Mohammad Shahadat and Andersson, Karl},
  journal={Procedia Computer Science},
  volume={238},
  pages={377--384},
  year={2024},
  publisher={Elsevier}
}

@inproceedings{islam2023sentigold,
  title={Sentigold: A large bangla gold standard multi-domain sentiment analysis dataset and its evaluation},
  author={Islam, Md Ekramul and Chowdhury, Labib and Khan, Faisal Ahamed and Hossain, Shazzad and Hossain, Md Sourave and Rashid, Mohammad Mamun Or and Mohammed, Nabeel and Amin, Mohammad Ruhul},
  booktitle={Proceedings of the 29th ACM SIGKDD Conference on Knowledge Discovery and Data Mining},
  pages={4207--4218},
  year={2023}
}

@article{paul2025analyzing,
  title={Analyzing Emotions in Bangla Social Media Comments Using Machine Learning and LIME},
  author={Paul, Bidyarthi and Rahman, SM and Biswas, Dipta and Hasan, Md Ziaul and Hossain, Md Zahid},
  journal={arXiv preprint arXiv:2506.10154},
  year={2025}
}

@inproceedings{paul2024improving,
  title={Improving Bangla Regional Dialect Detection Using BERT, LLMs, and XAI},
  author={Paul, Bidyarthi and Preotee, Faika Fairuj and Sarker, Shuvashis and Muhammad, Tashreef},
  booktitle={2024 IEEE International Conference on Computing, Applications and Systems (COMPAS)},
  pages={1--6},
  year={2024},
  organization={IEEE}
}

@article{paul2025ancholik,
  title={ANCHOLIK-NER: A Benchmark Dataset for Bangla Regional Named Entity Recognition},
  author={Paul, Bidyarthi and Preotee, Faika Fairuj and Sarker, Shuvashis and Refat, Shamim Rahim and Islam, Shifat and Muhammad, Tashreef and Hoque, Mohammad Ashraful and Manzoor, Shahriar},
  journal={arXiv preprint arXiv:2502.11198},
  year={2025}
}

@article{sultana2025onubad,
  title={ONUBAD: A comprehensive dataset for automated conversion of Bangla regional dialects into standard Bengali dialect},
  author={Sultana, Nusrat and Yasmin, Rumana and Mallik, Bijon and Uddin, Mohammad Shorif},
  journal={Data in Brief},
  volume={58},
  pages={111276},
  year={2025},
  publisher={Elsevier}
}






\end{document}